\documentclass[10pt,twocolumn,letterpaper]{article}

\usepackage{3dv}
\usepackage{times}
\usepackage{epsfig}
\usepackage{graphicx}
\usepackage{amsmath}
\usepackage{amssymb}
\usepackage{float}

\usepackage{latexsym}
\usepackage{amsfonts}
\usepackage{color}
\usepackage{comment}
\usepackage[abs]{overpic}
\usepackage{array}
\usepackage{multirow}
\usepackage{cite}
\usepackage[boxed]{algorithm2e}
\usepackage{caption}
\usepackage{subcaption}
\usepackage{array}
\usepackage[normalem]{ulem}
\usepackage{authblk}

\threedvfinalcopy

\def\threedvPaperID{125} 

\ifthreedvfinal\fi
\begin{document}

  \title{3D Face Reconstruction by Learning from Synthetic Data}

\author{ 
 Elad Richardson\textsuperscript{*} \qquad Matan Sela\textsuperscript{*} \qquad Ron Kimmel \\ Department of Computer Science, Technion - Israel Institute of Technology\\
    {\tt\small \{eladrich,matansel,ron\}@cs.technion.ac.il}
    }
  \maketitle

\begin{abstract}
Fast and robust three-dimensional reconstruction of facial geometric
 structure from a single image is a challenging task with numerous applications.
Here, we introduce a learning-based approach for reconstructing a three-dimensional face from a single image.
Recent face recovery methods rely on accurate localization of
 key characteristic points.
In contrast, the proposed approach is based on a
 Convolutional-Neural-Network (CNN) which extracts the face geometry
 directly from its image.
Although such deep architectures outperform other models
 in complex computer vision problems, training them properly requires
 a large dataset of annotated examples.
In the case of three-dimensional faces, currently, there are no large volume
 data sets, while acquiring such big-data is a tedious task.
As an alternative, we propose to generate random, yet nearly photo-realistic,
 facial images for which the geometric form is known.
The suggested model successfully recovers facial shapes from real images,
 even for faces with extreme expressions and under various lighting conditions.

\end{abstract}

  \section{Introduction}
  \let\thefootnote\relax\footnote{*Equal contribution}
Extracting the geometry of a surface embedded in an image is one of
 the fundamental challenges in the field of computer vision.
Generally speaking, the problem is ill-posed as different textured surfaces
 can be realized as the same image on a camera sensor.
Hence, when dealing with a certain class of shapes, it is common to
 employ prior knowledge for simplifying the problem.
In fact, reconstruction methods often differ in the way
 they utilize prior knowledge.

The unique challenge in face reconstruction is that although faces are,
 roughly speaking, very similar, there are local discrepancies
 between faces of different people.
Capturing these local differences is crucial for accurate facial geometry
 reconstruction.
A common approach for recovering face geometries is to apply an optimization
  procedure in which both geometry and lighting conditions are iteratively
  updated to best match the given image.
As an example, the analysis-by-synthesis approach introduced
 in~\cite{blanz1999morphable} alternates between rendering the reconstructed face
 and updating the representation and illumination parameters.
There, the prior comes in the form of the solution space, which is represented
 by a linear regression model built from a few hundred scans of facial geometries.
Although this approach can achieve highly accurate results, it requires
 manual initialization and focuses on faces with natural expressions.

  \begin{figure}[t]

    \centering
    \begin{subfigure}[b]{0.48\textwidth}
      \includegraphics[width=\textwidth]{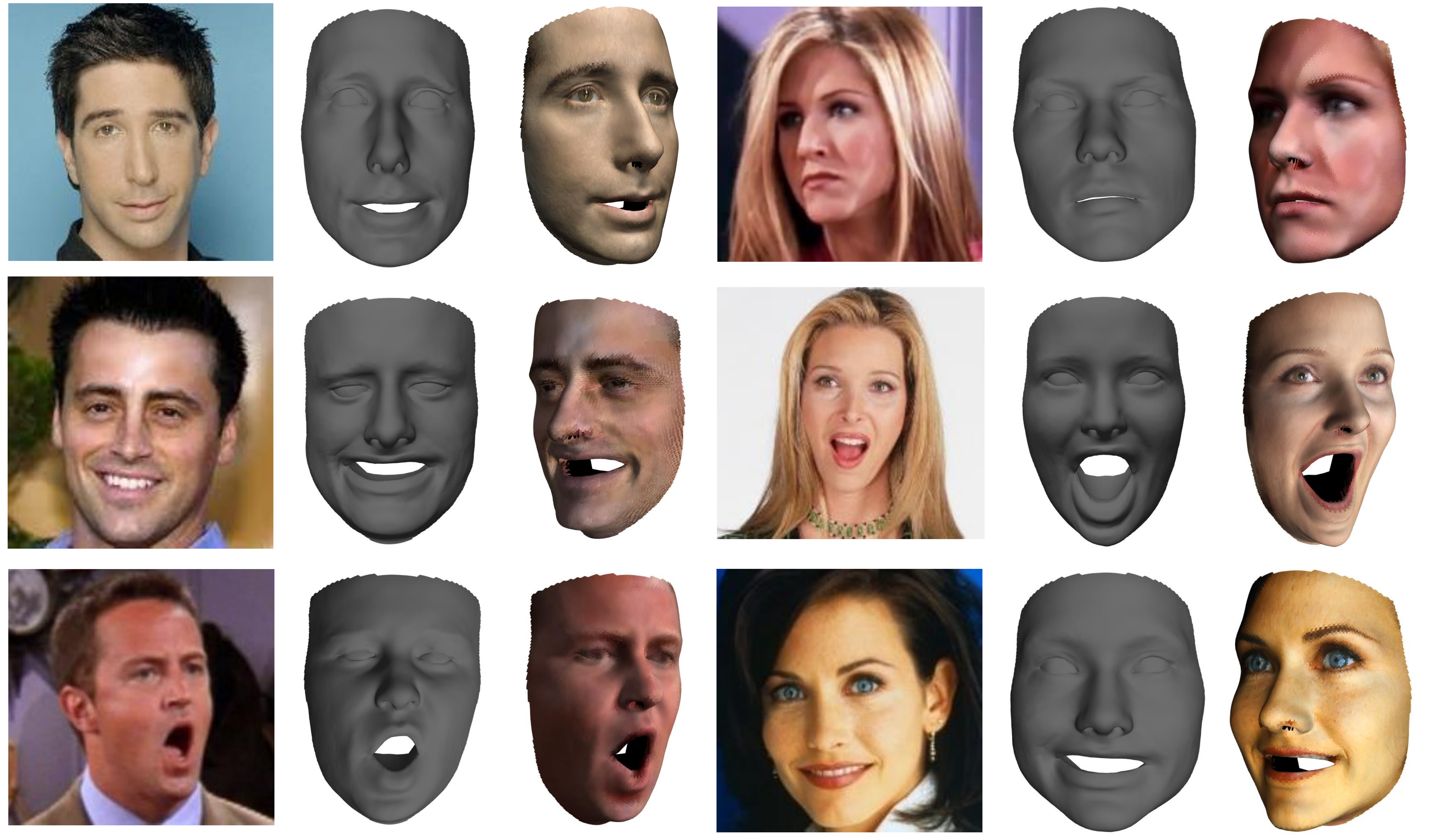}
    \end{subfigure}
    \caption{Reconstruction samples from our network.
     From left to right: a given facial image, a shading image of the reconstructed
     geometry representation predicted by our network,
     a textured reconstruction shown from a different viewing angle
     and under different lighting conditions.
    }
    \label{fig:res1}
\end{figure}
Alternatively, recent methods~\cite{aldrian2010linear,dou2014robust,liu2015cascaded}
 use a sparse set of 2D landmarks to directly infer the face geometry.
The usage of landmarks makes these algorithms more robust to lighting and albedo
 changes, as they are not directly dependent on the image.
However, these algorithms strongly depend on the landmark detection accuracy
 and cannot recover fine details which are not spanned by the sparse set of landmarks.
Here, we propose to train a neural network for directly reconstructing the
 geometric structure of a face embedded in a given image.
The benefits of using a neural network in this context are clear.
First, a network reconstructs the geometry based on the image as a whole.
Second, it can implicitly model different rendering methods, and third,
  it can be incorporated into real-time face reconstruction systems.

Convolutional neural networks have recently revolutionized computer vision
 research and applications with promising performance when applied to
 image classification and regression problems.
However, thousands of annotated samples are usually required for training
 a network which generalizes well.
In the case of three-dimensional faces, no suitable large dataset is currently
 available.
To that end, we propose to train our network with synthetic examples of
 photo-realistic facial images.
The faces we randomly draw from our model vary extensively in their pose,
 expression, race and gender.
Moreover, the images we synthesize have different illumination
  conditions, reflectance and background.
The proposed synthetic dataset construction allows us to introduce a
 computational model that can handle real images of faces in the wild.
That is, by training with artificial data that we synthesize, we can
 reconstruct surfaces of faces from natural images.

Our main contributions are as follows:
\begin{itemize}
 \item We introduce a novel method for generating a dataset of
     nearly photo-realistic facial images with known geometries.
 \item For the first time, a network is successfully trained to recover
          facial geometries from a single given image, based on synthetic examples.
 \item We exploit the fact that the space of facial textures and geometries
        can be captured by a low dimensional linear space.
        It is used in synthesizing the images and their corresponding
        geometries, and then in the reconstruction phase, where the
        output is a small number of coefficients in that space.
  \end{itemize}

\section{Related Efforts}
\label{sec:related}
Various approaches have been proposed for tackling the inherently ill-posed
 problem of  facial geometry reconstruction from a single image.
In~\cite{blanz1999morphable}, Vetter and Blanz observed that both the geometric
 structure and the texture of human faces can be approximated as a linear
  combination of carefully designed vectors.
For constructing this linear model, dubbed as the 3D Morphable Model (3DMM),
 they scanned a few hundred subjects, found a dense registration between them,
 and applied a principal component analysis on the corresponding scans.
For computing a plausible representation of a face in a given image,
  Vetter and Blanz proposed an analysis-by-synthesis approach which
  alternates between rendering a reconstruction and refining the geometry,
  texture, and illumination parameters according to the differences between
  the given image and the rendered one.

Alternatively, in~\cite{kemelmacher20113d}, Kemelmacher-Shlizerman and
 Basri proposed to recover the geometry using a shape from shading approach,
 where a single 3D reference face is manually placed in alignment with a given
 input facial image in order to constrain the problem.
A low-dimensional representation of reflectance based on spherical
 harmonics~\cite{basri03} is assumed, and the lighting coefficients are
 recovered based on the normals and albedo of the aligned reference face.
Then, the depth image and the albedo of the face are recovered based on
 the difference between the input image and the reconstructed one,
 generated with the recovered illumination parameters.
Note that both this method and the previous one explicitly employ the
 image acquisition model as part of their optimization and rely on an
 accurate initialization.

As landmark detection algorithms have become faster and
 robust~\cite{kazemi2014one,zhang2014facial,ren2014face,zhou2013extensive}
 it is now common to use them for facial reconstruction.
Some methods use these detection algorithms to automate the reconstruction
 process of existing methods.
As an example, in~\cite{breuer2008automatic}, feature points are used for
 rigidly aligning a 3D face model with the image.
The 3D face is then refined by deploying the method of
 Vetter and Blanz~\cite{blanz1999morphable}.
Still, automated initialization usually does not produce the same level of accuracy as manual ones, as shown in~\cite{piotraschke2016automated}.
Other methods~\cite{aldrian2010linear,dou2014robust,liu2015cascaded} choose
 to reconstruct the geometry solely from the detected landmarks.
These methods allow fast reconstructions and are usually more robust to
 lighting and albedo variations.
However, landmarks still incorporate only sparse information about the
 geometrical structure of the face, thus limiting the quality of the results.
This is sufficient for some tasks, such as in~\cite{zhu2015high},
 while other require further refinement as applied in~\cite{roth2016adaptive}
 for facial reconstruction from multiple images.

Recently, the 3DMM model was integrated with convolutional neural
 networks~\cite{zhu2015face,jourabloo2016large} for recovering pose variations
  of faces in images.
The networks were trained by augmentations of existing datasets of face
 alignment challenges.
These approaches suggest that given a suitable dataset, a CNN can
 successfully recover the morphable model coefficients from an image.

Training a network with synthetic data for shape recovery purposes was
 also suggested by Li \textit{et al.} in~\cite{li2015joint}.
There, a network was trained for extracting joint embedding of images and their
  corresponding 3D rigid objects.
The synthetic images were generated from the manually collected ShapeNet
 dataset~\cite{chang2015shapenet}.
The ShapeNet dataset was also recently used for the 3D object reconstruction
 method in~\cite{choy20163d}.
Here, we extend the concept of training from synthetic data by automatically
 generating a large dataset of facial images labeled with corresponding
 3DMM representations.

  \begin{figure*}[htbp]
    \begin{center}
      \begin{overpic}[width=1.0\textwidth]{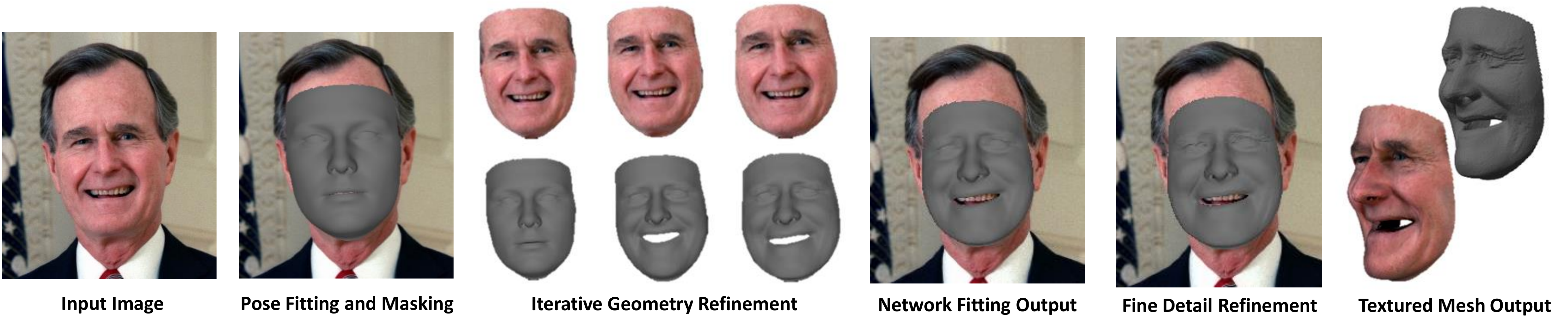}
      \end{overpic}
    \end{center}
    \caption{The pipeline of the proposed algorithm.
    The method starts with an arbitrary facial image
    and returns a detailed geometric facial reconstruction.
    }
    \label{fig:pipeline}
  \end{figure*}

  \section{Overview}\label{sec:overview}

An example of the proposed pipeline for recovering facial geometry
  from a single given image is visualized in Figure~\ref{fig:pipeline}.
  First, the input image is masked according to the projection of a generic 3D facial model on the image.
  The facial model is posed using the automatic alignment algorithm from ~\cite{kazemi2014one}.
The masked image is then propagated multiple times
 through the network, iteratively updating the geometry representation.
This network is trained on synthetic images
 of textured three-dimensional faces as shown in Section~\ref{sec:data}.
These images are rendered from geometric and photometric structures drawn
 from the 3DMM model as described in Section~\ref{sec:3dmm}.
The network architecture and training method is detailed in Section~\ref{sec:train}.
Finally, as an optional step, we demonstrate how the resulting facial structure
 can be further refined by a shape from shading algorithm in Section~\ref{sec:sfs}.


\section{The Face Model}
\label{sec:3dmm}
In order to model different plausible geometries, the 3D Morphable Model is used.
As in the framework proposed by Chu \textit{et al.} in~\cite{chu20143d},
 the representation is decomposed into identity and expression bases.
The identity basis $A_{id}$ is a compact principal component basis introduced
 in \cite{blanz1999morphable} and the expression basis $A_{exp}$ is a set
 of blendshapes, where each vector represents an offset from neutral facial
 identity to a specific expression.
The geometry of a given subject is linearly represented as
  \begin{equation}
    S=\mu_{S}+A_{\small\mbox{id}}\alpha_{\small\mbox{id}}
    +A_{\small\mbox{exp}}\alpha_{\small\mbox{exp}}.
  \end{equation}
Here, $\mu_{S}$ denotes the average shape.
  $\alpha_{\small\mbox{id}}$ and $\alpha_{\small\mbox{exp}}$ are the
   corresponding coefficient vectors of identity and expression, respectively.
As the proposed model is obviously restricted to geometries and
 expressions spanned by the proposed basis, examples need to be collected
 with care.
We used the reconstructions from the Bosphorus
 dataset~\cite{savran2008bosphorus},
 augmented with some in-house data, and aligned using a non-rigid
  ICP~\cite{Weise2009} procedure.
For spanning neutral geometric facial structures, we construct an identity
 basis with  200 vector elements, augmented by 84 vectors of
 various expressions.

The 3DMM model also represents the texture of the face as a linear
  combination of vectors.
This model is constructed similarly to the model of identity geometries by
  computing the principal components basis of the registered scanned faces.
Each texture is then represented as
\begin{equation}
    T=\mu_{T}+A_{T}\alpha_{T},
\end{equation}
  where $\mu_{T}$ is the average texture, $A_{T}$ is the texture
   basis and $\alpha_{T}$  is the coefficients vector.
For the texture model, we used $200$ basis elements.
In practice, using a significantly smaller number of basis elements
 for both geometry and texture produces comparable results.

\section{Learning Framework}
  \label{sec:train}
At the center of our pipeline lies the network,
  which predicts the face geometry from the image.
Next, we describe how this network operates.

\subsection{Iterative Formulation}
  \label{subsec:train-ief}

\begin{figure}[b]
    \centering
    \begin{subfigure}[b]{0.5\textwidth}
      \includegraphics[width=\textwidth]{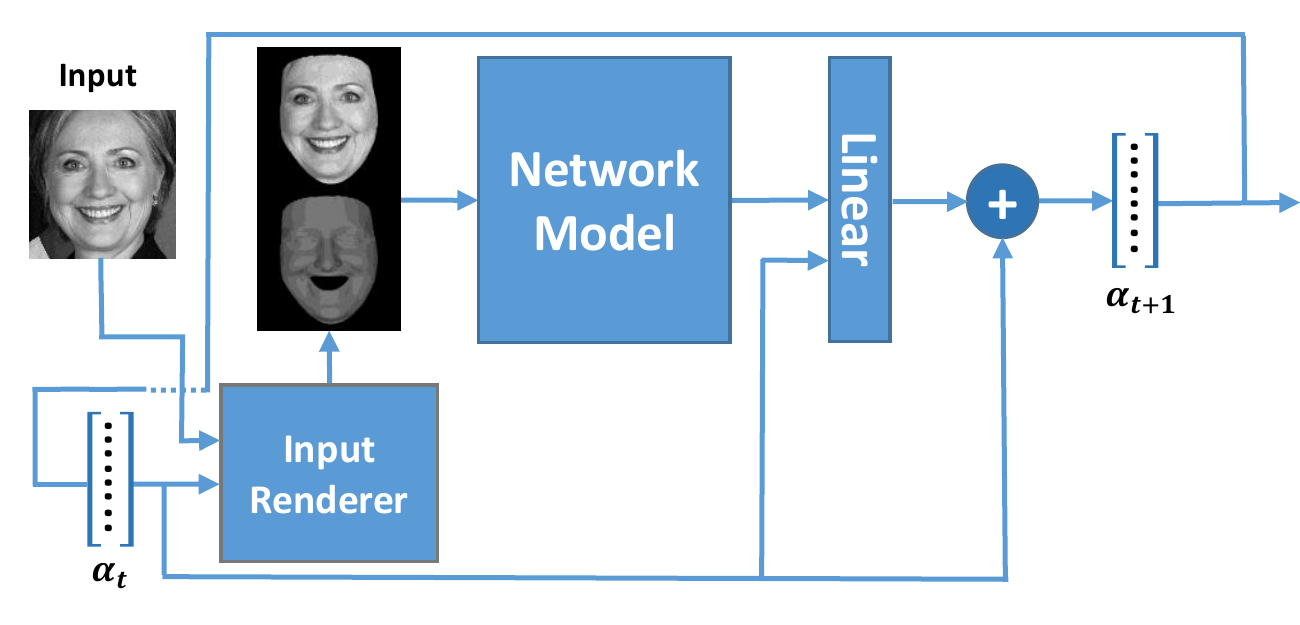}
    \end{subfigure}
    \caption{Iterative architecture. In every iteration inputs are generated based on the the previous prediction and propagated through the network. A linear layer then determines the geometry coefficient update.}
    \label{fig:net}
\end{figure}

In its basic formulation a Convolutional-Neural-Network (CNN)
 is a feed-forward system, predicting the outputs from the inputs
 in a single pass.
However, the power of one-shot algorithms is somewhat limited,
 as unlike iterative algorithms they cannot correct wrong predictions.
Inspired by~\cite{carreira2015human}, we run our network iteratively.
The core idea of iterative error feedback (IEF) is to use a secondary
 input channel to represent the previous network's output as an image.
The network can then be trained to correct the previous prediction based
 on both the original input and the secondary channel.
Representing the geometry coefficients as an image can be done in
 several different ways.
Here, we choose to represent it as a shading image from a frontal light
 source, where the geometry is aligned using an initial pose derived using~\cite{kazemi2014one}.
The IEF architecture is presented in Figure~\ref{fig:net}, alongside an
 example of the input channel.
Note that unlike~\cite{carreira2015human} we also add a linear layer
 which takes as input both the current network's output and the
  previous prediction, producing the actual correction.

During evaluation, we initialize the geometry using the average shape,
 setting $\alpha_{0}=0$.
This geometry is then used to create the shading image, as well as to
 mask the input image from the background.
At every iteration, a new geometry vector, $\alpha_{t}$, is predicted
 and used to update the shading image and the masking of the input image.
The procedure repeats $\sim$3 times improving the masking and the reconstruction iteratively.
The purpose of masking the face in the input image is to simplify the
 data generation process.
That way we only need to accurately synthesize the face itself.

\subsection{Training Criterion}
An important choice for the learning framework is the loss criterion.
A trivial choice in our case would be to use Mean Square Error (MSE)
  between the output representation vector and the ground-truth vector.
However, that formulation does not take into account how the different coefficients
  affect the geometry.
Instead, the criterion is defined as the MSE between the geometries themselves:
\begin{equation}
    L\left(x,y\right)=\left\Vert \left[A_{\small\mbox{id}}|A_{\small\mbox{exp}}\right]x
    -\left[A_{\small\mbox{id}}|A_{\small\mbox{exp}}\right]y\right\Vert _{2}^{2},
\end{equation}
  where $x$ is the output of the network, and $y$ is the known geometry.
  As this loss is differentiable it can be easily incorporated
   into the training procedure.
Note that all geometries are aligned, which justifies measuring error
  between  corresponding vertices.
Using this geometry MSE allows the network to take into account how changes
 in the coefficients would affect the final reconstruction error,
  and results in better convergence.

\subsection{Network Architecture}

The proposed network architecture is based on ResNet~\cite{he2015deep},
 winner of the ILSVRC 2015 classification task.
The architecture is detailed in Figure~\ref{fig:archi}, where the input
 is of size 200x200x2.
The additional linear layer mentioned in Section~\ref{subsec:train-ief}
 has an input of size 568, and an output of size 284.
The network was optimized using the ADAM method
  \cite{kingma2014adam}  with a fixed learning rate.
   \begin{figure}[h]
    \centering
    \begin{subfigure}[b]{0.45\textwidth}
      \begin{center}
        \begin{tabular}{c|c|c}
        \hline
        layer & stride & output size\tabularnewline
        \hline
        7x7, 32 & 2 & 100x100x32\tabularnewline
        b{[}32{]} x 2 & 1 & 100x100x32\tabularnewline
        b{[}64{]} x 2 & 2 & 50x50x64\tabularnewline
        b{[}128{]} x 2 & 2 & 25x25x128\tabularnewline
        b{[}256{]} x 2 & 2 & 13x13x256\tabularnewline
        3x3, 284 & 1 & 13x13x284\tabularnewline
        averaging & - & 284\tabularnewline
        \hline
        \end{tabular}
    \end{center}
    \end{subfigure}
    \caption{Network architecture. b[n] denotes a building block,
     as in the ImageNet architecture of~\cite{he2015deep},
     with n output maps. Padding is used in all layers.}
    \label{fig:archi}
  \end{figure}

\section{Data Generation}
  \label{sec:data}
An essential part of every learning algorithm is the training dataset.
Without a good training set of sufficient size, the learning model
  would simply fail to characterize the data.
For CNNs, where models with millions of parameters are commonly used,
  the generalization problem becomes even more complex, and
  one must often rely on large scale datasets to train a model.
However, existing datasets of 3D faces usually consist of just
  a few hundreds of subjects, making them unsuitable
  for deep learning tasks.
Ideally, one could capture enough 3D faces and use them to train
 a CNN model similarly to the suggested method.
However, scanning millions of faces with an accurate depth sensor is currently impractical.
Alternatively, one could take a set of 2D images and apply a
  reconstruction algorithm to generate the geometric
  representation.
Still, such an approach would limit our reconstruction capability
  to that of the specific geometry reconstruction algorithm we use,
  and, in practice, we will merely learn the 
  potential shortcomings of that algorithm.
Instead, we propose to directly generate different geometries using
  a morphable model.
Each such geometry is then rendered under random lighting conditions
  and projected onto the image plane.
This gives us a set of images for which the ground truth geometry
 is known.
In practice, generating a 3D face using 3DMM requires drawing a
 random normally distributed coefficients vector for the model
 described in Section~\ref{sec:3dmm}.
 
   \begin{figure}[]
    \centering
    \begin{subfigure}[b]{0.45\textwidth}
      \includegraphics[width=\textwidth]{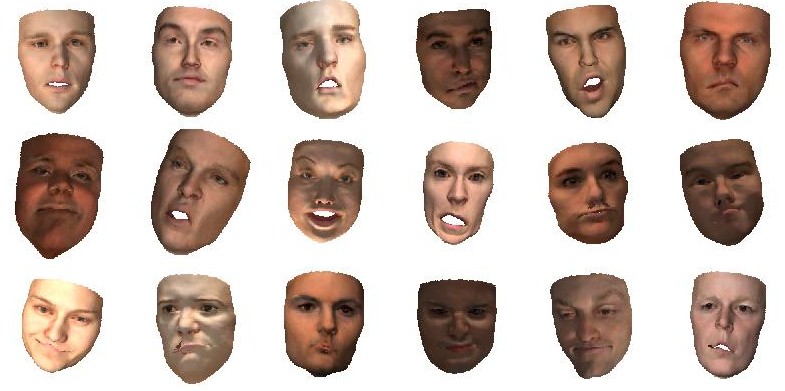}
    \end{subfigure}
    \caption{Samples of the synthetic images generated by
		    the proposed model.}
    \label{fig:gen}
  \end{figure}

\subsection{Rendering The Geometries}\label{sec:render}

Given a textured 3D geometry, a synthetic image can be rendered.
First, random shading needs to be drawn to create photo-realistic results.
We use Phong reflectance~\cite{phong1975illumination} for modeling the shading.
In our formulation the shininess constant was fixed at 10, while the the ambient,
 diffuse and specular constants were normally drawn around mean values
 of $\left[0.5,0.7,0.05\right]$ respectively.
The light direction was drawn from a uniform distribution, covering all possible frontal angles.

The shaded geometry is then projected onto the image plane,
  with a parallel weak perspective projection, as
  \begin{equation}
    \left[\begin{array}{c}
    p_{x}\\
    p_{y}
  \end{array}\right]=\left[\begin{array}{ccc}
  f & 0 & 0\\
  0 & f & 0
\end{array}\right]\left[R|t\right]\left[\begin{array}{c}
P_{x}\\
P_{y}\\
P_{z}\\
1
\end{array}\right].
\end{equation}
Rotation, translation and scaling parameters are all normally sampled and the
 mean value is set to form a front-facing centered face.
Geometries sampled from the proposed model are shown in Figure~\ref{fig:gen}.

\subsection{Iterative Data Simulation}
\label{sec:IDS}
The network operates in an iterative error feedback fashion,
 as detailed in Section~\ref{subsec:train-ief}.
This means that during evaluation the network always receives two inputs.
The first one being the cropped input image, while the second one is the shading image,
 representing the current geometry estimation.
The network then predicts a correction to the geometry estimation.
For training our network, a dataset of such samples needs to be generated.
That is, each dataset sample needs to include a cropped input image, some geometry estimation,
 and the ground-truth geometry  as a label.

\begin{figure}[h]
  \centering
  \begin{subfigure}[b]{0.45\textwidth}        \includegraphics[width=\textwidth]{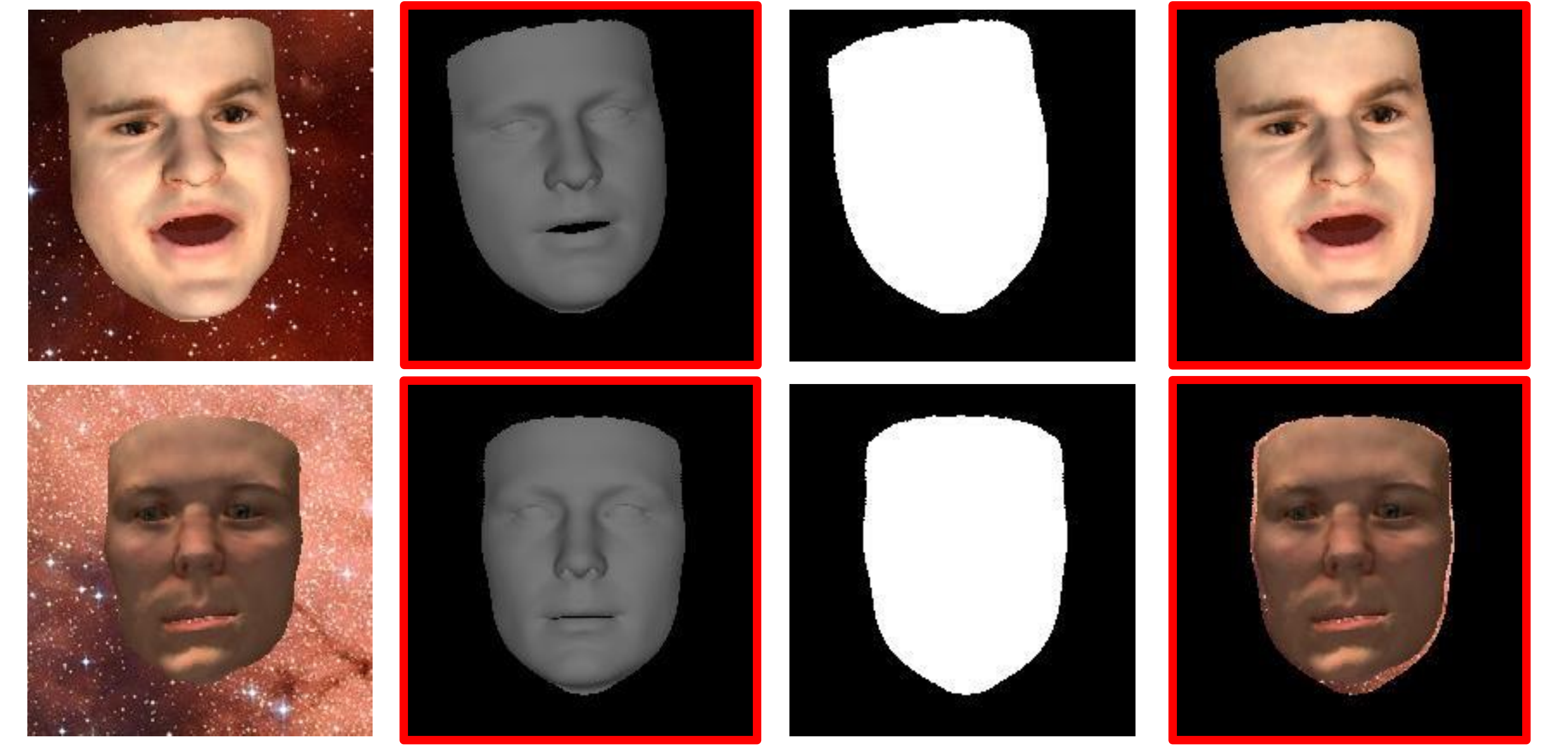}
  \end{subfigure}
  \caption{Training data sample generation.
  From left to right: Generated facial image. Generated secondary geometry rendered as a shading image. Binary mask of the shading image. Cropped facial image. The training pairs of the network are the images marked in red. }
  \label{fig:masking}
\end{figure}

The actual generation is done as follows.
First, we generate a random 3D face by drawing a ground-truth geometry, $\alpha_{gt}$, and a texture.
The 3D face is then rendered as explained in the previous section.
We then proceed by drawing another geometry, $\alpha_{t}$, as our current geometry estimation. $\alpha_{t}$ is then used to generate the shading image and to mask the rendered facial image.
See Figure~\ref{fig:masking} for generation examples.
Note that the same projection is used for both geometries.
The sampling of $\alpha_{t}$ needs to simulate the process of refining the initial
 geometry towards the real one.
This is being achieved by uniformly sampling it between some random geometry and $\alpha_{gt}$.
That way, our dataset includes samples in various distances from $\alpha_{gt}$.

%

\section{Extracting Fine Details}
\label{sec:sfs}
One of the limitations of using 3DMM to model the space of plausible geometries
 is that the results are constrained to the affine subspace spanned by the 3DMM vectors.
Consequently, 3DMM cannot model fine details such as the wrinkles around the eyes.
To solve this, we employ the real-time shape-from-shading (SFS) algorithm of Or-El \textit{et al.} \cite{or2015rgbd}. The predicted geometry is used as input to the algorithm, which then refines the reconstruction.
Note that the predicted geometry needs to be accurate at a coarse scale
 for the SFS refinement to work.
Experiments show that this formulation can successfully reconstruct fine
 details when used as a post-processing step.

\begin{figure}[h]
  \centering
  \begin{subfigure}[b]{0.48\textwidth}
    \includegraphics[width=\textwidth]{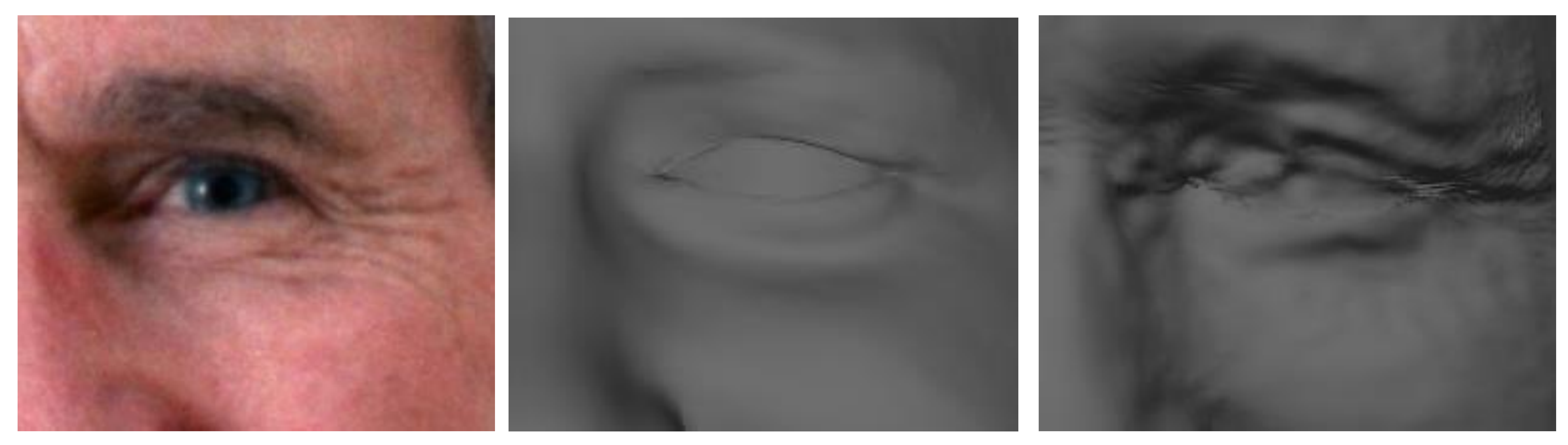}
  \end{subfigure}
  \caption{Shape from shading refinement. From left to right: the input image, the network-based reconstruction, and the same reconstruction after shape from shading.}
  \label{fig:sfs}
\end{figure}

\section{Results}

\begin{figure*}[t]
  \begin{center}
    \begin{overpic}[width=1.0\textwidth]{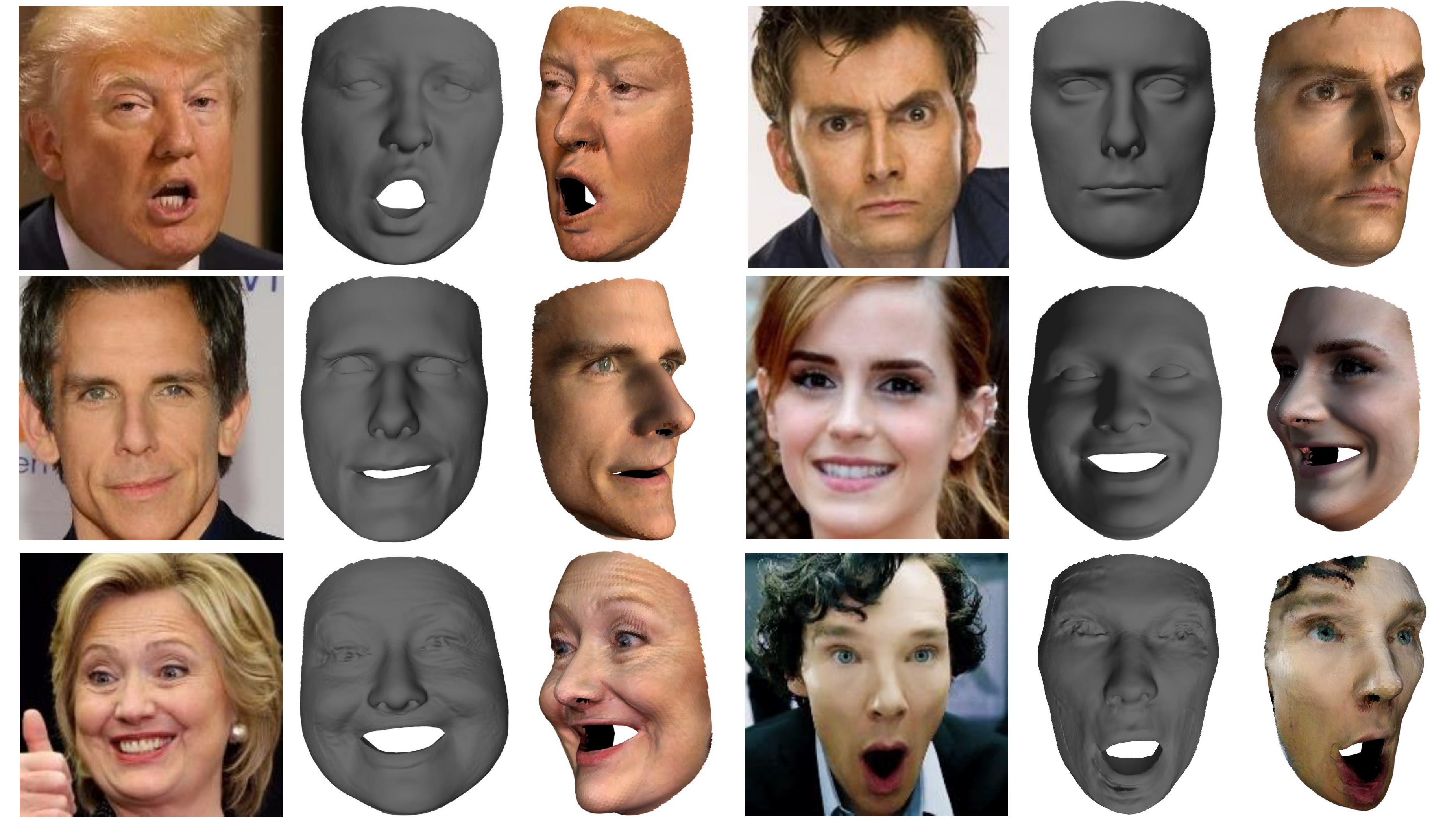}
    \end{overpic}
  \end{center}
  \caption{Results of the proposed algorithm.
  The first two rows visualize reconstructions without a shape from shading phase,
  while the last row presents results with the refinement step.}
  \label{fig:results-all}
\end{figure*}

\begin{figure}[h]
  \centering
  \begin{subfigure}[b]{0.45\textwidth}
    \includegraphics[width=\textwidth]{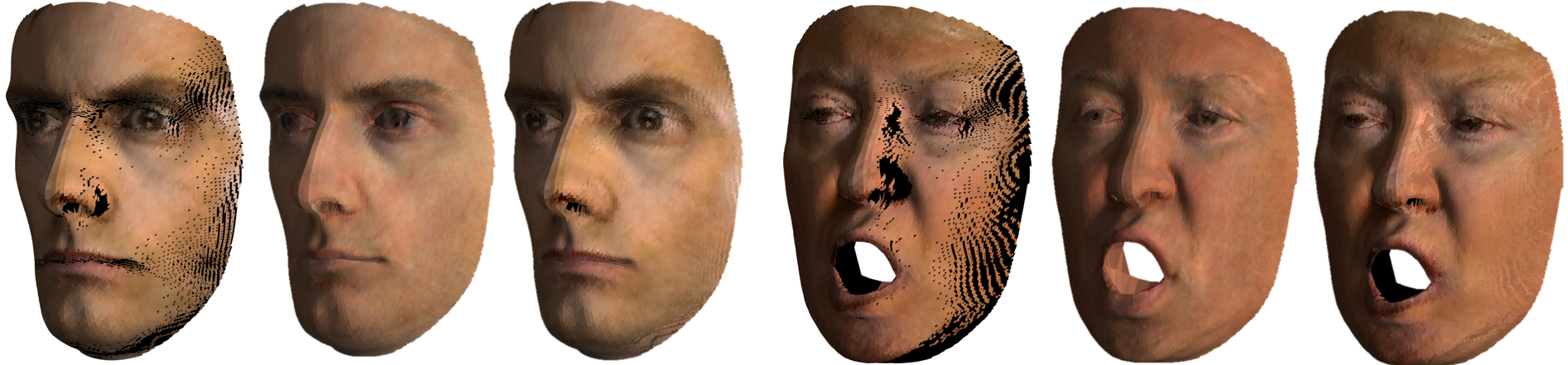}
  \end{subfigure}
  \caption{Texturing process. From left to right: The initial texture with missing
   values derived from the image, the texture retrieved from the linear model,
    and the combined one which is used for presenting the results in this paper.}
  \label{fig:texture}
\end{figure}

To show the generalization capability of our method we evaluate it on image of real
 faces \textit{in-the-wild}.
For a visual analysis of the reconstructions, we recover the texture according to the image
 and render the colored face from different viewing direction and under various
 lighting conditions.
Note, that some regions of the face are naturally occluded.
This prohibits a full recovery of the texture.
In order to fill these gaps, we first project the recovered texture onto the linear
 texture model, and set the missing values according to the reconstructed texture.
The texturing process is visualized in Figure~\ref{fig:texture}.

Different results of the proposed method are shown in
 Figures~\ref{fig:res1} and~\ref{fig:results-all}.
As demonstrated, the method successfully reconstructs \textit{in-the-wild}
 faces under different illumination conditions, viewpoints and expressions.
The effect of the SFS phase can be scrutinized in Figure~\ref{fig:sfs}, where
 zoomed-in images of the reconstructions are shown.

\begin{figure}[b]
  \centering
  \begin{subfigure}[b]{0.48\textwidth}
    \includegraphics[width=\textwidth]{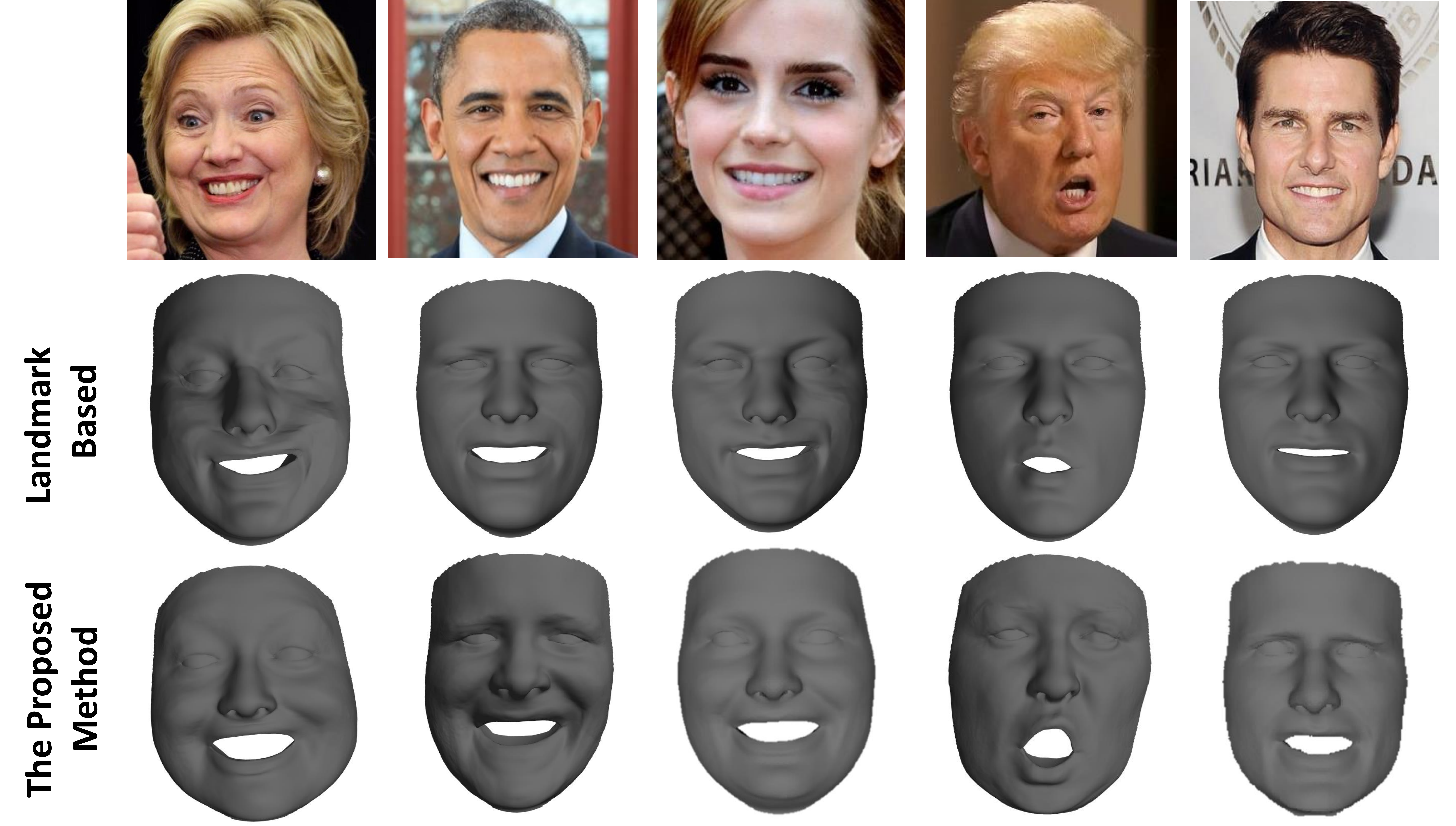}
  \end{subfigure}
  \caption{Method Comparisons (without SFS).
   The first row shows the input image.
  The second row shows reconstruction based only on landmark points.
  The third row shows the reconstruction results of the proposed methods.}
  \label{fig:compare}
\end{figure}

As mentioned in Section~\ref{sec:related}, landmark detection algorithms
 are utilized by different methods for 3D face reconstruction.
One common technique is to estimate a set of 3DMM parameters by minimizing
 the $L_2$ difference between the 2D image landmarks and the projected 3DMM landmarks.
In Figure~\ref{fig:compare} we compare the reconstruction achieved directly based
 on 68 image landmarks to the one generated by the proposed network.
As demonstrated, the proposed method successfully reconstructs more details,
 such as the shape of the nose and the wrinkles around the mouth,
 which are not spanned by the sparse set of landmarks.
%

\begin{figure}[h]
  \centering
  \begin{subfigure}[b]{0.48\textwidth}
    \includegraphics[width=\textwidth]{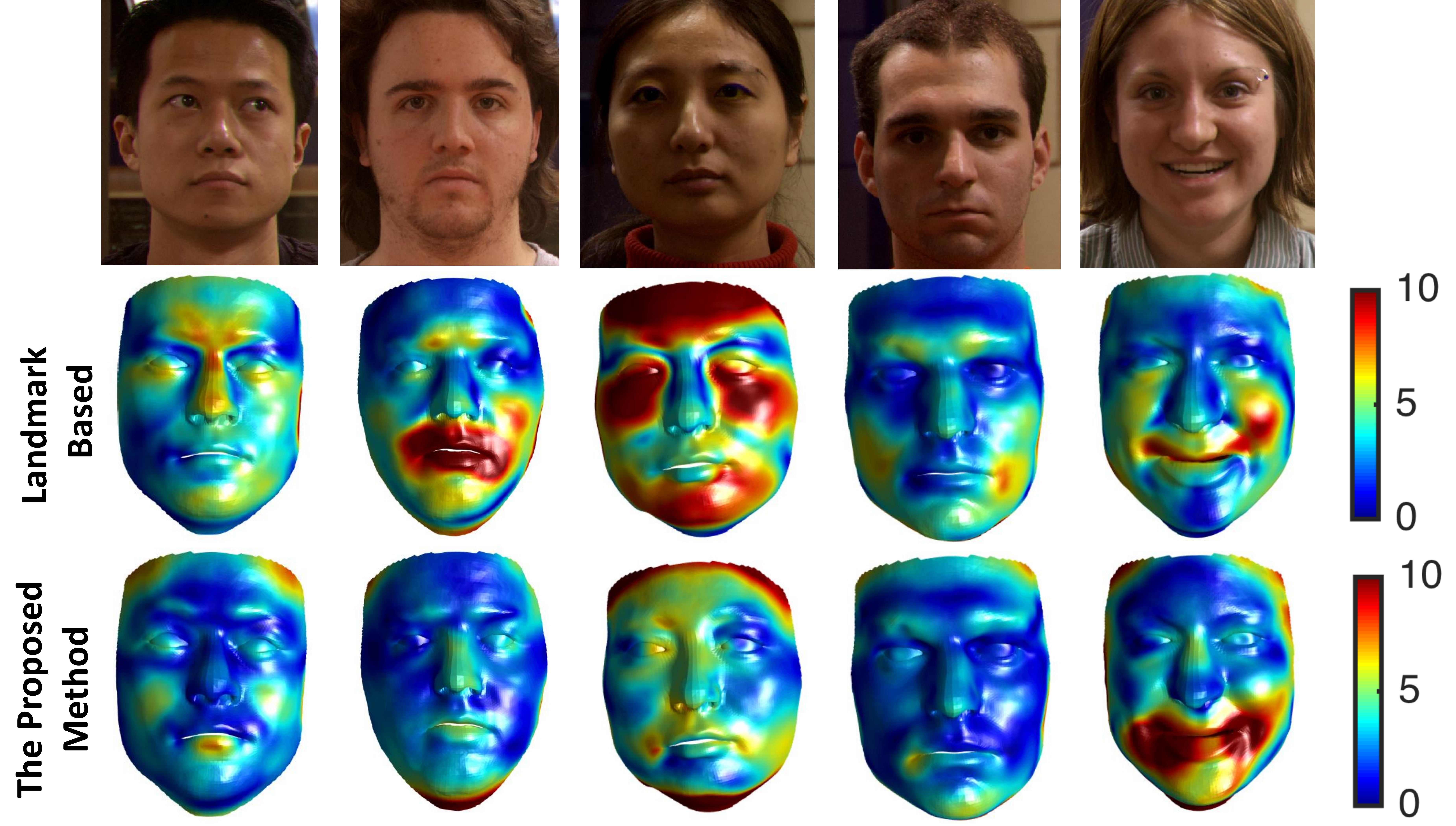}
  \end{subfigure}
  \caption{Quantitative analysis of the proposed method vs. landmark point based reconstruction. Each face is colored according to the pointwise Euclidean error (in millimeters) of the reconstruction.}
  \label{fig:quant}
\end{figure}

A quantitative comparison between the proposed method and a landmark based reconstruction
 is given in Figure~\ref{fig:quant}.
For evaluating the accuracy of the reconstruction we registered our average shape to
 several 3D scans from the FRGC-v2.0  dataset using a non-rigid registration,
  and used the warped template as our ground truth.
Then, we aligned each reconstruction with the ground truth by the optimal
 similarity transform, and computed the pointwise Euclidean distance between
 corresponding points on the two shapes.
As demonstrated, in these cases, the network recovers the subtle details of the
 face more accurately.

\section{Discussion}
The proposed approach was shown to recover relatively accurate facial geometry
 from a single image.
In our experiments, we explored other alternative configurations.
For example, we compared the usage of full-perspective projections versus weak
 perspective ones.
According to our findings, the latter outperformed the former.
We believe that the superiority of the weak perspective projection can be
 explained by the ambiguity in initializing the face mask parameters
 under a full-perspective projection.
That is, a narrow face may correspond to either a round geometry close to the
 camera or a narrow one viewed from a distance.

There are numerous possible options for extending the proposed method.
For example, as the suggested network does not require explicit representation of the
 rendering model, more accurate rendering techniques can be easily incorporated
 to create more realistic synthetic images without changing the learning framework.
One such option is to use a ray casting shader which imitates more precisely the
 image acquisition process.
Intuitively, the more similar the synthetic data is to real \textit{in-the-wild} images
 the better the network's result should be.
Another possible direction is to incorporate face landmarks as another input into
 the suggested algorithm, utilizing the robustness of landmark detection.
Finally, as all components of the suggested pipeline can be implemented to run in
 real-time on a GPU, the proposed method can possibly be used for real-time face
 reconstruction from video.
The iterative formulation of the network would allow us to solve every frame
 based on the previous one without modifying the architecture.


\subsection{Limitations}
\begin{figure}[h]
  \centering
  \begin{subfigure}[b]{0.45\textwidth}
    \includegraphics[width=\textwidth]{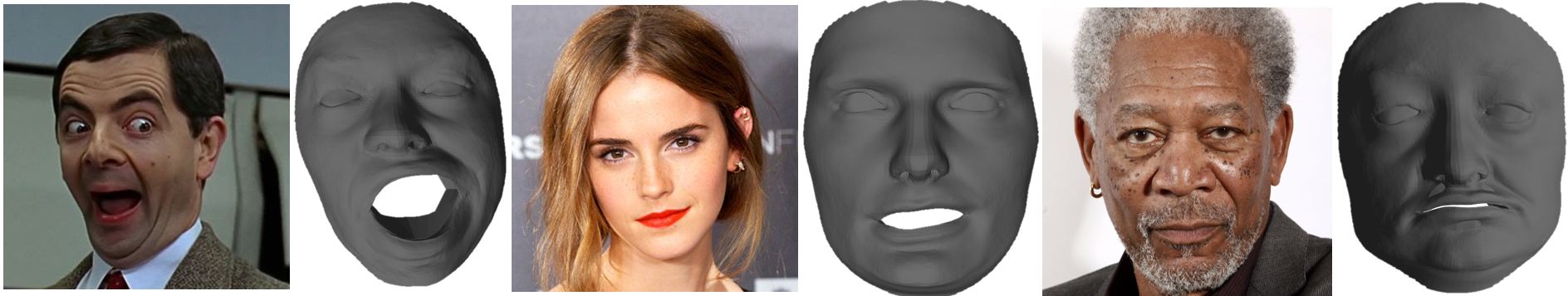}
  \end{subfigure}
  \caption{Limitations of the reconstruction.}
  \label{fig:fail-recon}
\end{figure}

Although, in many cases, the suggested algorithm performs well, we
 encountered some scenarios which prohibited the network from recovering the geometry.
Firstly, since the 3DMM model was constructed by scanning faces from specific
 ethnic origins, the proposed approach sometimes fails to fully recover images
 of faces which are not spanned by the scanned faces.
Secondly, we found that the network occasionally makes mistakes when faced with
 facial attributes which were not present in the synthetic data.
For example, thick lips with lipstick could be mistaken for the interior of the mouth,
 while beards could be mistaken for parts of the background, as shown
  in Figure \ref{fig:fail-recon}.
Both problems can be solved by incorporating a richer model such as the one proposed
 in \cite{booth3d} and using better, and more accurate, rendering techniques.

\section{Conclusion}
We proposed an efficient algorithm for recovering facial geometries from a single image.
The proposed approach reconstructs the face based on the image as a whole.
For this, we employ an iterative Convolutional-Neural-Network
 trained with synthetic data.
As an optional detail refinement step, a shape-from-shading algorithm was applied.
We evaluated the method by recovering facial surfaces from various pictures
 of faces {\it in-the-wild}.
As demonstrated, the proposed approach successfully generalizes from our synthetic
 data and can efficiently and accurately handle a large variety of expressions
  and different illumination conditions.

\subsubsection*{Acknowledgments}

The research leading to these results has received funding from the 
European Research Council under European Union’s Seventh Framework Programme, 
ERC Grant agreement no. 267414 as well as
European Research Council (ERC) under the European Union’s
Horizon 2020 research and innovation programme (grant agreement No 664800).
The authors would like to thank Mr. Roy Or-El for providing the source code of the shape from shading algorithm.
{\small
\bibliographystyle{ieee}
\bibliography{references}
}

\end{document}